\documentclass[10pt]{article} 

\usepackage{fullpage}
\usepackage{enumitem}
\usepackage[page]{appendix}
\usepackage{amssymb}
\usepackage{mathrsfs} 
\usepackage{natbib}
\usepackage{algorithm}
\usepackage{algpseudocode}

\usepackage{hyperref}
\usepackage{comment}
\usepackage{multirow}
\usepackage{amsthm}
\usepackage{mathtools}
\usepackage{epstopdf,xcolor}
\usepackage{bbm}
\usepackage[colorinlistoftodos,prependcaption]{todonotes}
\usepackage{dsfont}
\usepackage{framed}
\usepackage{pifont}
\usepackage{subcaption}
\usepackage{tikz, subcaption}
\usetikzlibrary{bayesnet, shapes, arrows, positioning}
\usepackage{authblk}
 \usepackage{listings}
 \usepackage{tabularx}

\title{\bf \LARGE Root Cause Attribution of Delivery Risks\\ via Causal Discovery with Reinforcement Learning\\~}

\author[]{Minheng Xiao$^1$\footnote{Authors are equally contributed.}}
\affil[]{
$^1$Department of Mathematics and Statistics, Boston University\\
$^2$Department of Integrated System Engineering, Ohio State Univeristy}

\begin{document}

\maketitle

\begin{abstract}
This paper presents a novel approach to root cause attribution of delivery risks within supply chains by integrating causal discovery with reinforcement learning. As supply chains become increasingly complex, traditional methods of root cause analysis struggle to capture the intricate interrelationships between various factors, often leading to spurious correlations and suboptimal decision-making. Our approach addresses these challenges by leveraging causal discovery to identify the true causal relationships between operational variables, and reinforcement learning to iteratively refine the causal graph. This method enables the accurate identification of key drivers of late deliveries, such as shipping mode and delivery status, and provides actionable insights for optimizing supply chain performance. We apply our approach to a real-world supply chain dataset, demonstrating its effectiveness in uncovering the underlying causes of delivery delays and offering strategies for mitigating these risks. The findings have significant implications for improving operational efficiency, customer satisfaction, and overall profitability within supply chains. \\


\end{abstract}

\section{Introduction}
In today’s increasingly complex and interconnected global economy, supply chain management has emerged as a critical factor in ensuring the success and competitiveness of businesses \citep{ketchen2008best}. Effective management of supply chains is essential for improving customer satisfaction, optimizing productivity, and maintaining a competitive edge in the market. As businesses navigate the challenges of globalization and evolving geopolitical landscapes, they must prioritize the monitoring and analysis of customer delivery performance. By closely tracking delivery processes and analyzing relevant metrics, companies can gain valuable insights into their operations, identify areas for improvement, and proactively address issues before they escalate into significant disruptions \citep{christopher2004building}. 

Supply chain disruptions can have profound economic impacts, leading to delays in delivery and production as well as increased costs \citep{chopra2004managing}. Understanding the root causes of these disruptions is crucial for mitigating their effects and ensuring the smooth functioning of supply chains. Traditionally, root cause analysis has relied on a combination of subject matter expertise and classical statistical methods to identify the underlying issues within supply chains \citep{heizer2020operations}. However, as supply chains grow in complexity, these traditional methods struggle to keep pace with the increasing number of potential root causes and their intricate interrelationships \citep{blanchard2021supply}. Moreover, conventional statistical approaches are often prone to identifying spurious correlations, which may lead to misguided conclusions and ineffective interventions. For example, a specific manufacturing plant might appear to be linked to frequent delays, or certain shipment times may seem correlated with late deliveries, but these correlations do not necessarily reflect true causal relationships \citep{neuberg2003causality}. Inaccurate attribution of causality can result in suboptimal decision-making, ultimately exacerbating supply chain inefficiencies \citep{glymour2019review}. To date, Most existing studies on delayed delivery risk have focused on the response to the risk after it occurs, ignoring how the risk arises and what is causal relationship between those risk factors.

To address these challenges, we propose a novel approach that integrates causal discovery combined with reinforcement learning methods~\citep{sutton1999reinforcement, yu2022risk, yu2023global, xiao2024policy, wang2024research} and causal strength to perform root cause attribution of delivery risks within supply chains. Causal discovery is a powerful tool that identifies the underlying causal relationships between variables based on observational data, thereby providing a more accurate and interpretable understanding of the factors that influence delivery outcomes \citep{peters2017elements}. By leveraging reinforcement learning, our approach not only identifies these causal relationships but also optimizes the process of constructing and refining the causal graph, ensuring that the most relevant and impactful factors are considered in the analysis \citep{zhu2019causal}.

Our method introduces several key innovations by integrating domain-specific knowledge into the causal discovery process, enabling more accurate identification of causal directions that align with real-world supply chain dynamics. The model inherently understands that certain relationships, such as the influence of product specifications on demand, do not work in reverse. This awareness allows our reinforcement learning framework to refine the causal graph continuously based on data feedback, enhancing the precision of root cause attribution over time \citep{bengio2013estimating}.

In supply chain management, understanding the causal relationships between operational variables is crucial for making informed decisions that prevent disruptions. Our method identifies whether factors like shipping mode, days for shipment, or order characteristics influence the likelihood of late deliveries, either directly or indirectly. This not only provides actionable insights for optimizing supply chain performance but also offers explainable framework by clarifying the underlying causes of delivery risks. Our approach thus serves as a robust solution to the challenges of root cause attribution in complex supply chain environments, making it easier for managers to implement targeted interventions and understand the reasoning behind them.

\textbf{Paper Organization} In Section \ref{sec:2}, we outline our proposed method for root causes of delivery risks. In Section \ref{sec:3}, we conduct an exploratory data analysis (EDA) to identify key patterns and relationships in the data and analyze the experimental results. In Section \ref{sec:4}
,  we discuss the potential business impact of our work. We conclude in Section \ref{sec:5} and discuss future work.

\subsection{Existing Works in Supply Chain on Delivery Risk}
Most of the existing studies on delayed delivery risk have concentrated on the countermeasures implemented after the risk has materialized. For example, some approaches focus on ensuring delivery by optimizing production processes \citep{biccer2017optimal, chen2017optimizing} or by enhancing delivery strategies \citep{bushuev2018delivery, shao2018production}. Nevertheless, these studies often overlook the factors that lead to the emergence of these risks in the first place.

Other studies have employed machine learning and deep learning techniques to predict delivery risks and optimize supply chain operations. For instance, some researchers have applied machine learning algorithms such as random forests, support vector machines, clustering with attention and neural networks to predict delivery delays by analyzing historical data and identifying patterns associated with late shipments \citep{chong2018predicting, ryu2019predicting, lim2019prediction,wang2019supply, bo2024attention}. These models have shown effectiveness in improving the accuracy of delivery time predictions, allowing companies to better anticipate delays and take preemptive actions. Despite their predictive power, these approaches often treat the delivery process as a black box, lacking the ability to explain the underlying causes of the risks, which limits their usefulness for making strategic decisions \citep{verma2019supply}.

In addition to predictive models, optimization techniques have been widely adopted to enhance supply chain resilience. For example, several studies have explored the use of optimization algorithms to adjust inventory levels, routing plans, and production schedules in response to predicted delays, aiming to minimize the impact of disruptions on the supply chain \citep{cruijssen2019enhancing, mourtzis2019real, mishra2020smart}. These methods typically focus on the immediate responses to delivery risks, such as rerouting shipments or increasing inventory buffers. While these strategies can mitigate the effects of delays, they do not address the root causes of the risks, leaving the underlying issues unresolved and potentially leading to recurring problems \citep{zhou2018supply}.

Furthermore, there has been significant interest in leveraging real-time data and Internet of Things (IoT) technologies to monitor and respond to delivery risks dynamically. Studies have demonstrated the potential of IoT-enabled systems to provide real-time visibility into the supply chain, allowing for more agile and responsive management of delivery risks \citep{lee2018internet, janjua2019real, lin2020real}. These systems use sensor data and real-time analytics to detect anomalies and trigger automated responses, such as adjusting delivery routes or alerting managers to potential delays. However, while these technologies enhance the ability to manage risks as they arise, they still do not offer insights into the causal mechanisms behind these risks, which are crucial for developing long-term solutions \citep{liu2020iot}.

While significant progress has been made in predicting and managing delivery risks using advanced technologies and optimization techniques, current research largely overlooks the underlying causes of these risks. Understanding what triggers delays and how these risks arise is essential for developing more effective and sustainable strategies to prevent disruptions in the supply chain. This gap highlights the need for approaches that not only predict delivery risks but also elucidate the causal relationships between different factors in the supply chain, enabling a more comprehensive understanding and better management of delivery risks.

\section{Mothedology \label{sec:2}}
\subsection{The proposed method}
We adopt the data-generating model as proposed by \citep{hoyer2008nonlinear, peters2014causal}, where each variable \( x_i \) corresponds to a node \( i \) in a \( d \)-node directed acyclic graph (DAG) \( \mathcal{G} \). The observed value of \( x_i \) is modeled as a function of its parent variables in the graph, plus an independent additive noise \( n_i \). Specifically,
\begin{align}
    x_i := f_i(\mathbf{x}_{\text{pa(i)}}) + n_i, \quad i = 1, 2, \ldots, d,
\end{align}
where \( f_i(\mathbf{x}_{\text{pa(i)}}) \) represents the set of parent variables \( x_j \) that have directed edges towards \( x_i \) in the graph, and the noise terms \( n_i \) are assumed to be jointly independent. We also assume causal minimality, which in this context implies that each function \( f_i \) is non-constant in any of its arguments, as noted by \citep{peters2014causal}.

\begin{figure}
    \centering
    \includegraphics[width=1.1\linewidth]{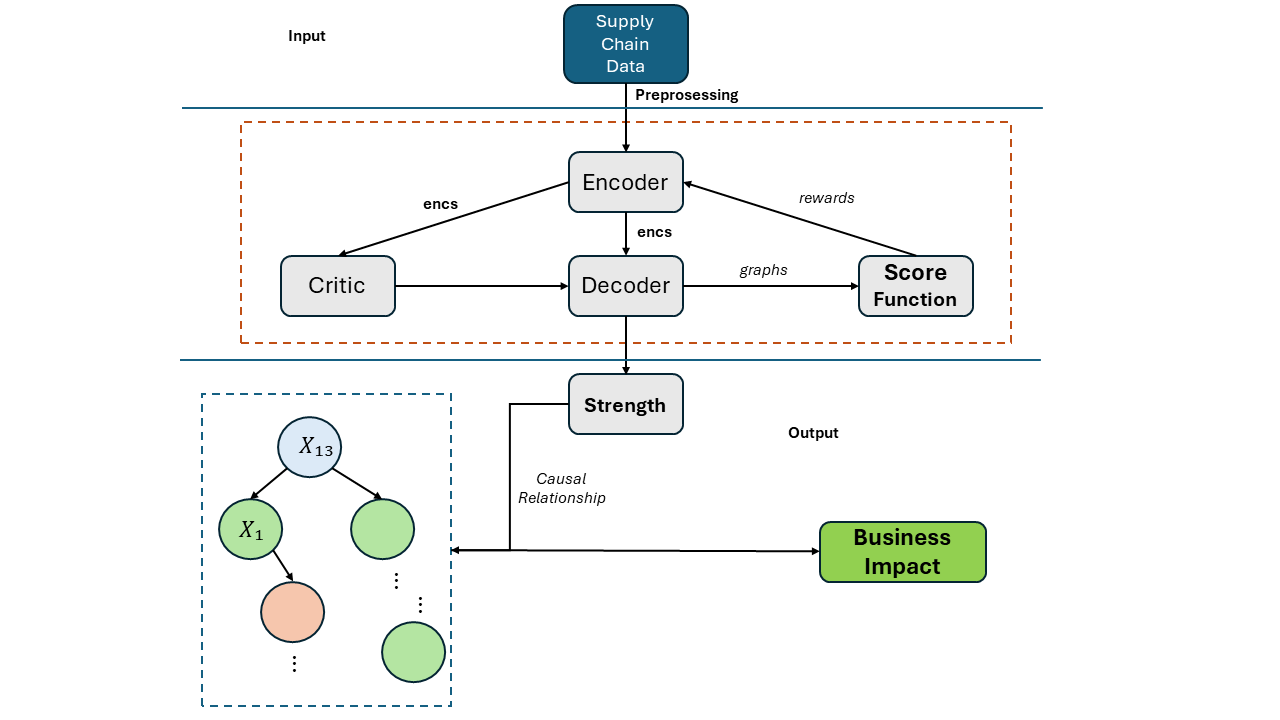}
    \caption{Proposed causal discovery approach with reinforcement learning for delivery risk}
    \label{fig:model}
\end{figure}

\begin{figure}[H]
    \hspace*{2cm} 
    \includegraphics[width=1\linewidth]{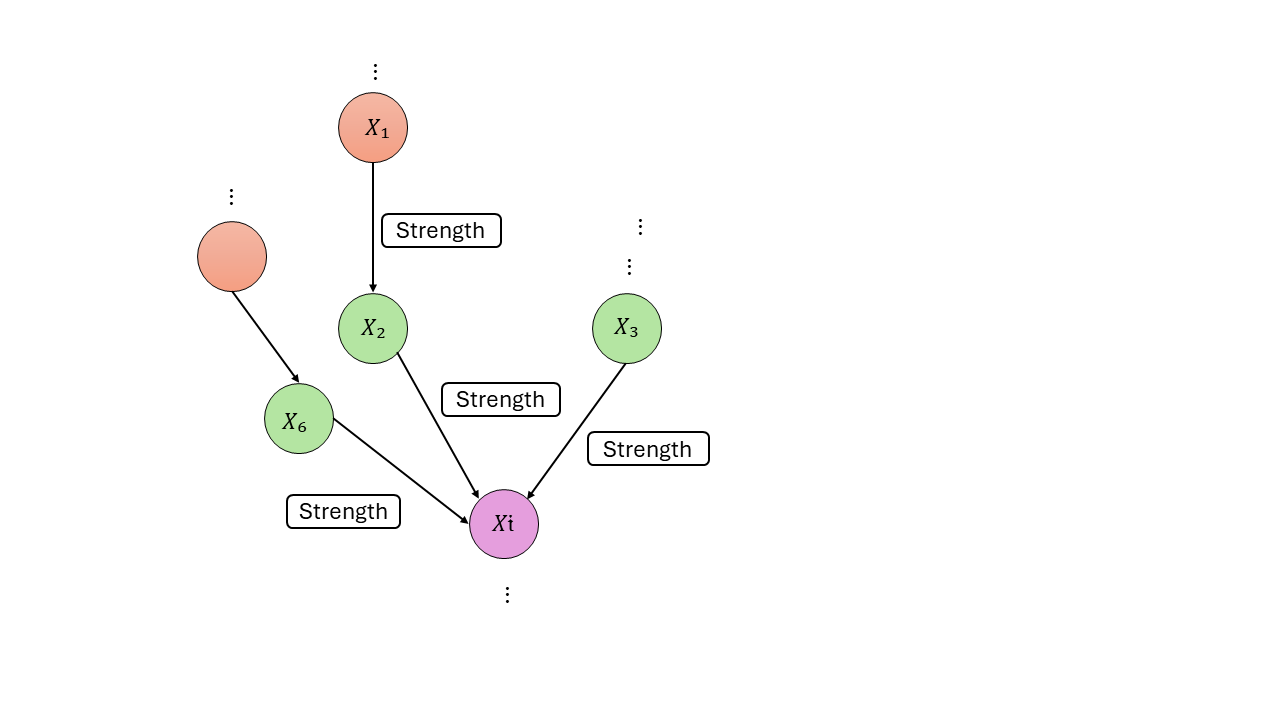}
    \caption{Causal structure}
    \label{fig:structure}
\end{figure}

As illustrated in Fig.\eqref{fig:model}, our proposed model consists of two parts: causal discovery via reinforcement learning and the calculation of inverse information entropy (IIE) causal strength \citep{mu2022inverse}. The model takes as input the observed dataset \( X = \{x_1, x_2, \ldots, x_i\} \), where \( x_i \) denotes the dimensions of the input observations, and produces as output a causal structure comprising a causal graph \( G \) and causal strengths. Fig.\eqref{fig:structure} presents the causal structure simulated by the observed data \( O = \{G, S\} \), where \( S \) indicates the strength of the causal relationships in graph \( G \).

Our encoder-decoder approach builds upon the work of \cite{zhu2019causal}. The model uses an encoder-decoder framework to generate a directed graph. The encoder, identical to the one in \cite{vaswani2017attention}, consists of six identical layers, each containing two sublayers. The first sublayer is a multi-headed self-attention mechanism, and the second is a fully connected feedforward network with positional encoding. The sublayers are connected using residual connections as described by \cite{he2016deep}. The output of each sublayer is normalized as follows:
\begin{align}
    \text{Layer Norm}(x + \text{Sublayer}(x)),
\end{align}
where \( \text{Sublayer}(x) \) represents the function implemented by the sublayer. To maintain consistency, all sublayer and embedding outputs in the model have the same dimension, \( d_{\text{model}} = 512 \). Considering the connections between different variables, we employ a single-layer decoder, defined as:
\begin{align}
\label{eq:gw}
    g(W_1, W_2, u) = u^\top \text{tanh}(W_1 \text{enc}_i + W_2 \text{enc}_j),
\end{align}
where \( W_1, W_2 \in \mathbb{R}^{d_h \times d_n} \) and \( u \in \mathbb{R}^{d_h \times 1} \) are trainable parameters, \( d_h \) is the number of hidden layers in the decoder, and \( d_n \) is the dimension of the encoder output. To generate the adjacency matrix, each element is passed through a sigmoid function and then sampled based on a Bernoulli distribution with probability \( \sigma(g) \):
\begin{align}
    \label{eq:ber}
M \sim \text{Ber}(\sigma(g)),
\end{align}

To prevent self-loops, the diagonal elements \( (i, i) \) of the adjacency matrix are set to zero. By iteratively inputting the encoder outputs of all variables, a complete directed graph's adjacency matrix is obtained. The scoring function uses the Bayesian Information Criterion (BIC), which is decomposable and allows for adjusting the penalty term. The BIC score for a graph \( \mathcal{G} \) is given by:
\begin{align}
    S_{\text{BIC}}(\mathcal{G}) = -2 \log p(X ; \hat{L} ; \mathcal{G}) + d_{L} \log m,
\end{align}
where \( \hat{L} \) is the maximum likelihood estimate, \( d_L \) is the number of parameters in \( L \), and \( m \) is the size of the dataset \( X \). To ensure that the generated graph is a DAG, the score function includes both a reward and a penalty term, incorporating two acyclic constraints:
\begin{align}
    \text{reward} = -\left[S_{\text{BIC}}(\mathcal{G}) + \lambda_{1} I(\mathcal{G} \notin \text{DAGs}) + \lambda_{2} h(A)\right],
\end{align}
where \( I(\cdot) \) is an indicator function, \( \lambda_1, \lambda_2 \geq 0 \) are hyperparameters, \( A \in \{0, 1\}^{d \times d} \), and \( h(A) \) is a function introduced by \cite{zheng2018dags}, which is non-negative and small for cyclic graphs. The binary adjacency matrix of a directed graph \( \mathcal{G} \) is acyclic if and only if:
\begin{align}
    h(A) = \operatorname{trace}\left(e^{A}\right) - d = 0,
\end{align}
where \( e^A \) is the matrix exponential of \( A \). It is evident that larger values of \( \lambda_1 \) and \( \lambda_2 \) increase the likelihood that a graph with a high reward is acyclic. Our goal is to maximize the reward across all possible directed graphs, which is equivalent to solving:
\begin{align}
\label{eq:maxreward}
    \min_{\mathcal{G}} \left[S_{\text{BIC}}(\mathcal{G}) + \lambda_1 I(\mathcal{G} \notin \text{DAGs}) + \lambda_2 h(A)\right].
\end{align}
The expected return during training can be expressed as:
\begin{align}
    J(\varphi \mid s) = \mathbb{E}_{A \sim \pi(\cdot \mid s)}\left\{-\left[S_{\text{BIC}}(\mathcal{G}) + \lambda_{1} I(\mathcal{G} \notin \text{DAGs}) + \lambda_{2} h(A)\right]\right\},
\end{align}
where \( \pi(\cdot \mid s) \) is the strategy and \( \varphi \) represents the neural network parameters for graph generation. During training, random samples are drawn from the observed dataset \( X \). The encoder output is fed into the critic, which is a simple two-layer feedforward neural network with a tanh activation function. The critic minimizes the mean squared error between predicted and actual rewards and penalties, and is trained using the Adam optimizer. Moving on to the IIE, building on \cite{mu2022inverse}, the causal strength is defined as:
\begin{align}
\label{eq:strength}
    T = \frac{1}{\left|S\left(p_{X_2}\right) - S\left(p_{X_1}\right)\right|},
\end{align}
where \( S(p_{x_1}) \) and \( S(p_{x_2}) \) represent the information entropies of variables \( x_1 \) and \( x_2 \), respectively. For a finite dataset, the entropy of the probability distribution for risk factors can be estimated using the entropy estimator as described in \citep{daniusis2012inferring,kraskov2004estimating}:
\begin{align}
    \hat{S}(X) = \psi(n) - \psi(1) + \frac{1}{n-1} \sum_{i}^{n-1} \log \left|x_{i+1} - x_{i}\right|,
\end{align}
where \( \psi(n) \) is the digamma function, and \( n \) represents the dimensionality of the dataset \( X \). Based on Eq. \eqref{eq:strength}, the IIE causal strength is computed using raw data. However, due to potential differences in dimensionality among variables in the raw data, the calculated causal strength might deviate from the actual value. Thus, the IIE causal strength for normalized data is given by:
\begin{align}
    T_{N} = \frac{1}{\left|S\left(p_{X_2, N}\right) - S\left(p_{X_1, N}\right)\right|}.
\end{align}

Eventually, We applied log transformation to the computed causal strengths to normalize their distribution and stabilize variance.. This transformation enhances the interpretability of the results, making it easier to compare the relative impact of different factors on delivery risks.
\begin{align}
\label{eq:finalstrength}
    \text{Causal Strength} = \log (T_N)
\end{align}

\textbf{Algorithm:}\\
Next, we propose the algorithm for our causal discovery via RL for root cause attribution on late delivery. As can be seen in Algorithm \ref{algo}, the proposed approach is as following:

\begin{algorithm}[H]
\caption{Causal Discovery with RL for Late Delivery Risk}
\label{algo}
\begin{algorithmic}[1]
    \State \textbf{Input:} Data pertaining to late delivery risk.
    \State \textbf{Step 1:} Preprocess the observed data and set hyperparameters, including the number of epochs, learning rate, and scoring function.
    \State \textbf{Step 2:} Feed the preprocessed data into the encoder, and pass the encoder outputs along with those from the critic to the decoder. The decoder performs calculations using Eq. \eqref{eq:gw} and samples according to Eq. \eqref{eq:ber}.
    \State \textbf{Step 3:} Rate the generated directed graph using Eq. \eqref{eq:ber}, and return rewards and penalties to the critic. Track the maximum reward using Eq. \eqref{eq:maxreward}.
    \State \textbf{Step 4:} Check if the preset number of iterations has been reached. If not, return to Step 2; otherwise, output the directed graph with the maximum score.
    \State \textbf{Step 5:} From the final directed graph, calculate causal strengths with log transformation \eqref{eq:finalstrength}, remove edges with a strength less than 0.1, and output the refined causal graph with the associated strengths.
    \State \textbf{Step 6:} Generate an explainable delivery risk report from the output graph, tailored for business stakeholders.
\end{algorithmic}
\end{algorithm}

\section{Experiments \label{sec:3}}
\subsection{Experimental data}
\subsubsection{DataCo Global}
We use a dataset of supply chains by the company DataCo Global for the analysis. This dataset is designed for analyzing and modeling various aspects of supply chain management provided by Kaggle. The dataset have $180520$ samples and $53$ features. Our aim is to find out the root cause of late delivery risk by analyzing the dataset by our causal discovery via reinforcement learning with strength caculation. 

For the preprocessing of our dataset, we decided to begin by manually filtering out some variables that are less relevant to our specific problem, such as “Order Date”, “Order City”, and similar attributes. This step helps us focus on the most pertinent data, reducing noise and simplifying our analysis. After the initial manual filtering, we proceeded to check for multicollinearity among the remaining features. Multicollinearity can distort our model's predictions, as highly correlated features may carry redundant information. To address this, we removed features that exhibited high correlations with others, ensuring that each remaining feature contributes unique information to the model. Next, we transformed the categorical variables into numerical formats by encoding them as integers. This step was crucial for enabling the effective use of these variables in machine learning models, which typically require numerical input. Finally, after these preprocessing steps, we selected 16 key features that we identified as most relevant to our problem. These features will serve as the input for our models, providing a balanced and optimized dataset for analysis and predictive modeling. The remaining features are listed in Table \ref{table:var} and description of variables to use is recorded in Appendix \ref{table:des}. 
\begin{table}[H]
\centering
\begin{tabular}{|c|c|c|c|}
\hline
\textbf{Denotation} & \textbf{Variable} & \textbf{Denotation} & \textbf{Variable} \\ \hline
\( X_1 \) & Type & \( X_9 \) & Latitude \\ \hline
\( X_2 \) & Days for shipping (real) & \( X_{10} \) & Longitude \\ \hline
\( X_3 \) & Days for shipment (scheduled) & \( X_{11} \) & Order Item Discount Rate \\ \hline
\( X_4 \) & Benefit per order & \( X_{12} \) & Order Item Product Price \\ \hline
\( X_5 \) & Sales per customer & \( X_{13} \) & Order Item Profit Ratio \\ \hline
\( X_6 \) & Delivery Status & \( X_{14} \) & Order Item Quantity \\ \hline
\( X_7 \) & Late\_delivery\_risk & \( X_{15} \) & Order Status \\ \hline
\( X_8 \) & Customer Segment & \( X_{16} \) & Shipping Mode \\ \hline
\end{tabular}
\caption{Denotations and Their Corresponding Variables}
\label{table:var}
\end{table}
In the next subsection, we will conduct a comprehensive exploratory data analysis (EDA) on the dataset. This analysis will involve examining the data in detail to uncover underlying patterns, relationships, and trends that may be present. Through visualization techniques and statistical summaries, we will identify key insights that can inform our modeling approach and enhance our understanding of the factors driving the outcomes within the dataset.
\subsubsection{Explanatory Data Analysis}

\begin{figure}[H]
    \centering
    \begin{tabular}{cc}
        \includegraphics[width=0.482\textwidth]{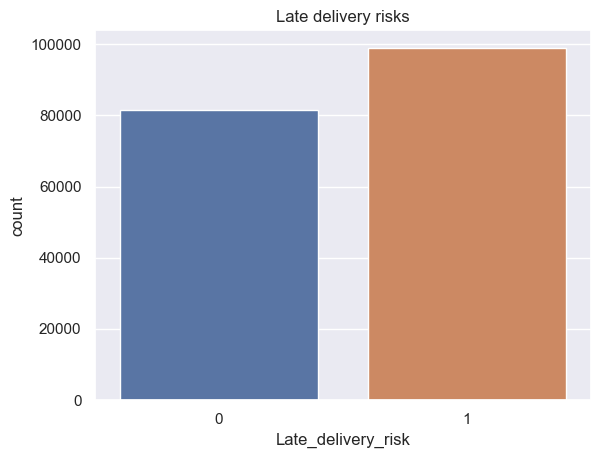} &
        \includegraphics[width=0.45\textwidth]{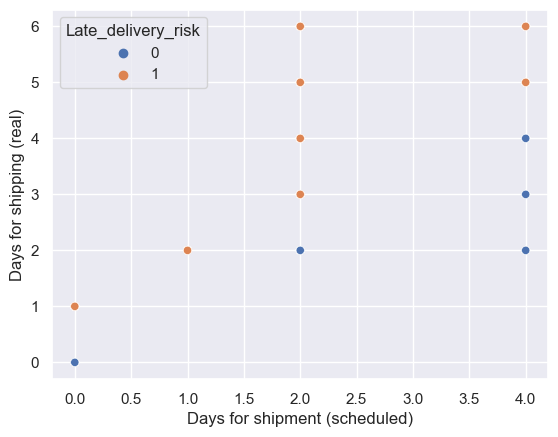} \\
        (a) Countplot of late delivery risks & (b) Late delivery risks by days of shipment \\
    \end{tabular}
    \caption{Analysis of late delivery risk}
    \label{fig:late}
\end{figure}

Fig. \eqref{fig:late} is an initial overview of the "Late\_delivery\_risk" ($X_7$) variable within the dataset, which is critical for our exploratory data analysis (EDA) before delving into causal discovery with reinforcement learning for root cause attribution. The first chart displays a countplot of late delivery risks, showing that instances of late deliveries (`1`) slightly outnumber on-time deliveries (`0`), suggesting that delays are a common issue in this dataset. The second chart, a scatter plot, explores the relationship between the actual days for shipping and the scheduled days for shipment, with different colors indicating the presence or absence of late delivery risk. The clustering of late deliveries around specific combinations of scheduled and actual shipment days hints at underlying patterns that could be driving these delays. Together, these visualizations highlight key areas where delays occur, providing a foundation for the causal analysis that will follow.

\begin{figure}[H]
    \centering
    \begin{tabular}{cc}
        \includegraphics[width=0.55\textwidth]{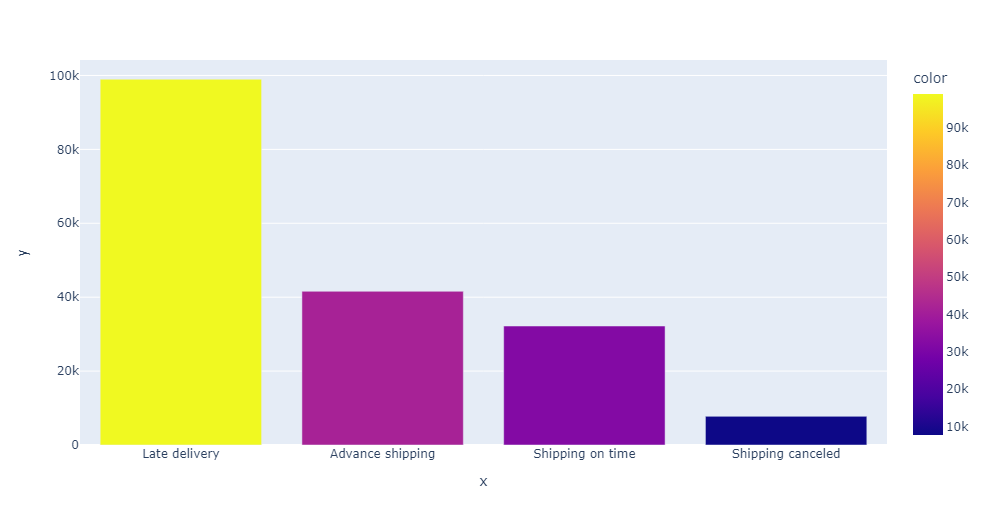} &
        \includegraphics[width=0.32\textwidth]{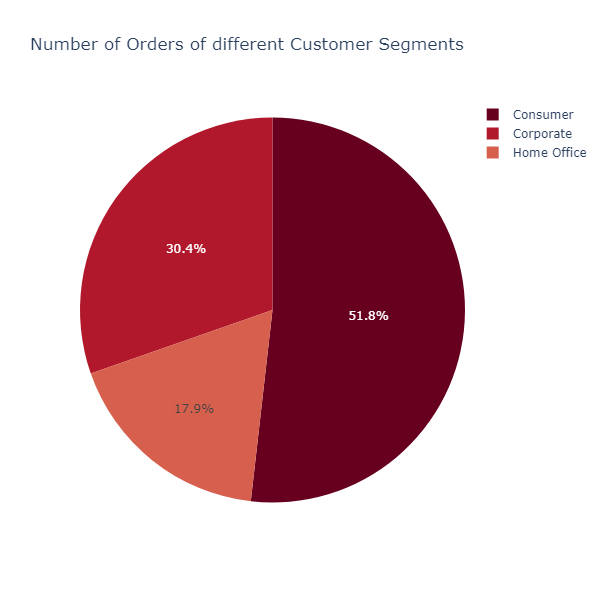} \\
        (a) Delivery Status & (b) Customer segements \\
    \end{tabular}
    \caption{Delivery status and customer segments}
    \label{fig:status}
\end{figure}

Fig. \eqref{fig:status} further contribute to our exploratory data analysis (EDA) by offering insights into delivery performance and customer segmentation, which are crucial for understanding the factors influencing late delivery risks. The bar chart on the left (a) illustrates the distribution of delivery statuses across the dataset. Notably, "Late delivery" is the most common outcome, significantly outnumbering "Advance shipping," "Shipping on time," and "Shipping canceled." This highlights a pervasive issue with delays, making it a key area of focus for our causal analysis. The prevalence of late deliveries suggests systemic issues that could be explored in greater depth, possibly linked to other variables in the dataset. We believe that Delivery Status ($X_6$) might cause the late delivery. 

The pie chart on the right (b) provides a breakdown of orders by customer segment ($X_8$). The majority of orders come from the "Consumer" segment (51.8\%), followed by "Corporate" (30.4\%), and "Home Office" (17.9\%). This distribution is important for our analysis, as it may reveal segment-specific factors contributing to late deliveries. Understanding the different behaviors and risks associated with each segment will be essential in accurately attributing the root causes of delivery delays. However, we have reservations about whether it will ultimately cause causation, since the segments may influence the days of shipping but may not impact the late delivery. 
\begin{figure}[H]
    \centering
    \includegraphics[width=0.95\linewidth]{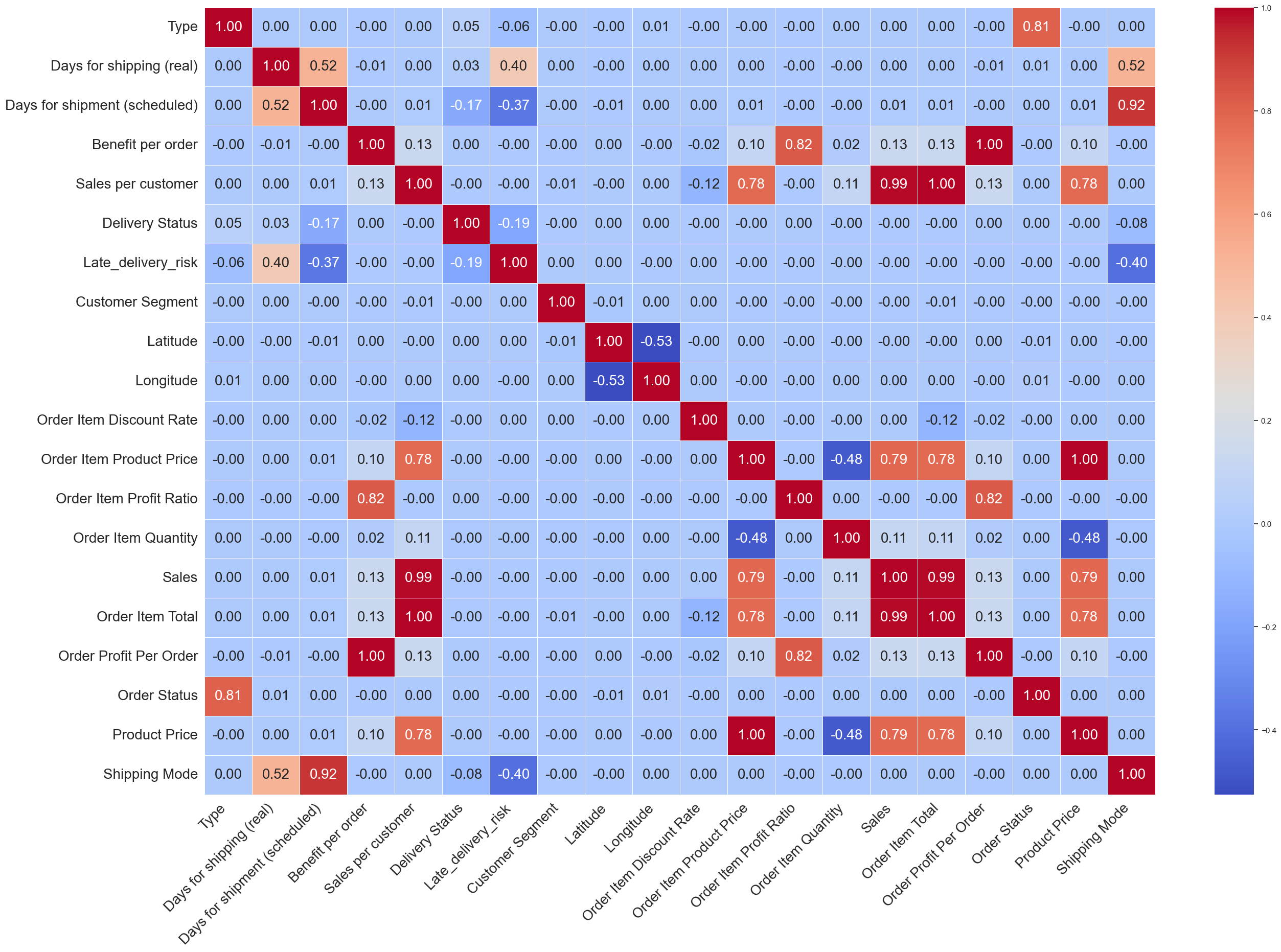}
    \caption{Correlation Matrix}
    \label{fig:corr}
\end{figure}
This heatmap visualization in Fig. \eqref{fig:corr} provides a correlation matrix for the key features in the dataset. Each cell in the matrix represents the correlation coefficient between two features, with the color intensity indicating the strength and direction of the correlation. Several noteworthy patterns emerge from this visualization. For instance, "Days for shipment (scheduled)" ($X_3$) and "Shipping Mode" ($X_{16}$) exhibit a strong positive correlation (0.92), suggesting that the shipping mode might significantly influence or be influenced by the scheduled shipment days. Similarly, "Sales per customer" ($X_5$), "Order Item Total", and "Sales" are highly correlated with each other, indicating potential redundancy in these features. 

Of particular interest to our study on late delivery risks, the "Late\_delivery\_risk" ($X_7$) variable shows a moderate negative correlation with "Days for shipment (scheduled)" ($X_3$) (-0.37) and a positive correlation with "Delivery Status" ($X_6$) (0.19). This suggests that shorter scheduled shipment times might be associated with a higher risk of late delivery, and that delivery status is indeed a significant factor linked to delays. Understanding these correlations is crucial as we prepare to apply causal discovery with reinforcement learning. By identifying and potentially removing highly correlated features, we can reduce multicollinearity, ensuring that the remaining features contribute unique and meaningful information to the causal model. This step is key to accurately attributing the root causes of late delivery risks within the dataset.
\subsection{Experimental analysis}
\begin{figure}[H]
    \centering
    \includegraphics[width=1.05\linewidth]{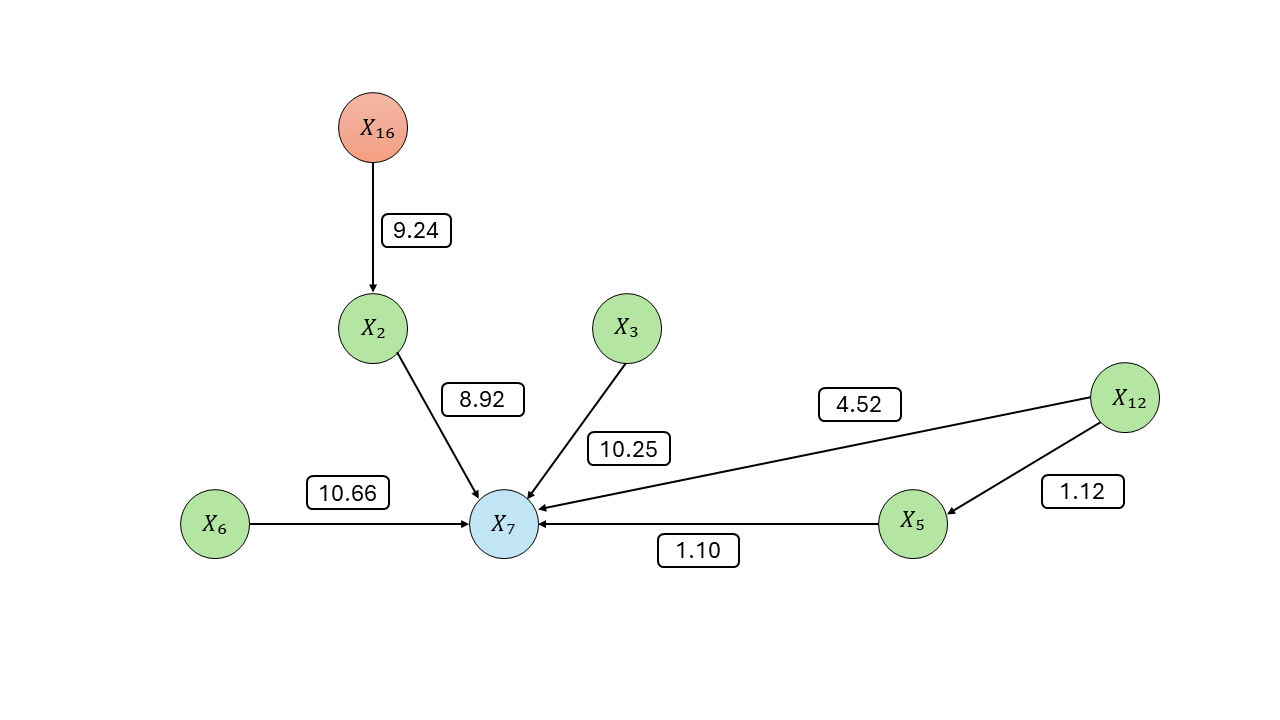}
    \caption{Causal structure for delivery risk}
    \label{fig:output1}
\end{figure}

The final causal graph in Fig. \eqref{fig:output1} illustrates the intricate relationships between various features and their impact on `Late\_delivery\_risk` (\(X_7\)), providing a detailed map of how different variables contribute to the likelihood of a late delivery. The shipping mode (\(X_{16}\)) emerges as a pivotal factor, exerting a strong direct influence on `Days for shipping (real)` (\(X_2\)) with a causal strength of 9.24. This, in turn, significantly affects both the `Days for shipment (scheduled)` (\(X_3\)) and directly influences the `Late\_delivery\_risk` (\(X_7\)), with strengths of 8.92 and 10.25, respectively. This indicates that the choice of shipping mode sets off a chain of effects that ultimately determine whether a delivery will be late.

Moreover, the `Delivery Status` (\(X_6\)) has a direct impact on `Late\_delivery\_risk`, reflected by a causal strength of 10.66, highlighting that certain statuses during the delivery process are strong indicators of potential delays. The interconnectedness of these nodes suggests that delays in actual shipping time not only directly heighten the risk of late delivery but also indirectly influence other variables, such as scheduled shipping times, thereby compounding the delay.

\begin{figure}[H]
    \centering
    \includegraphics[width=1\linewidth]{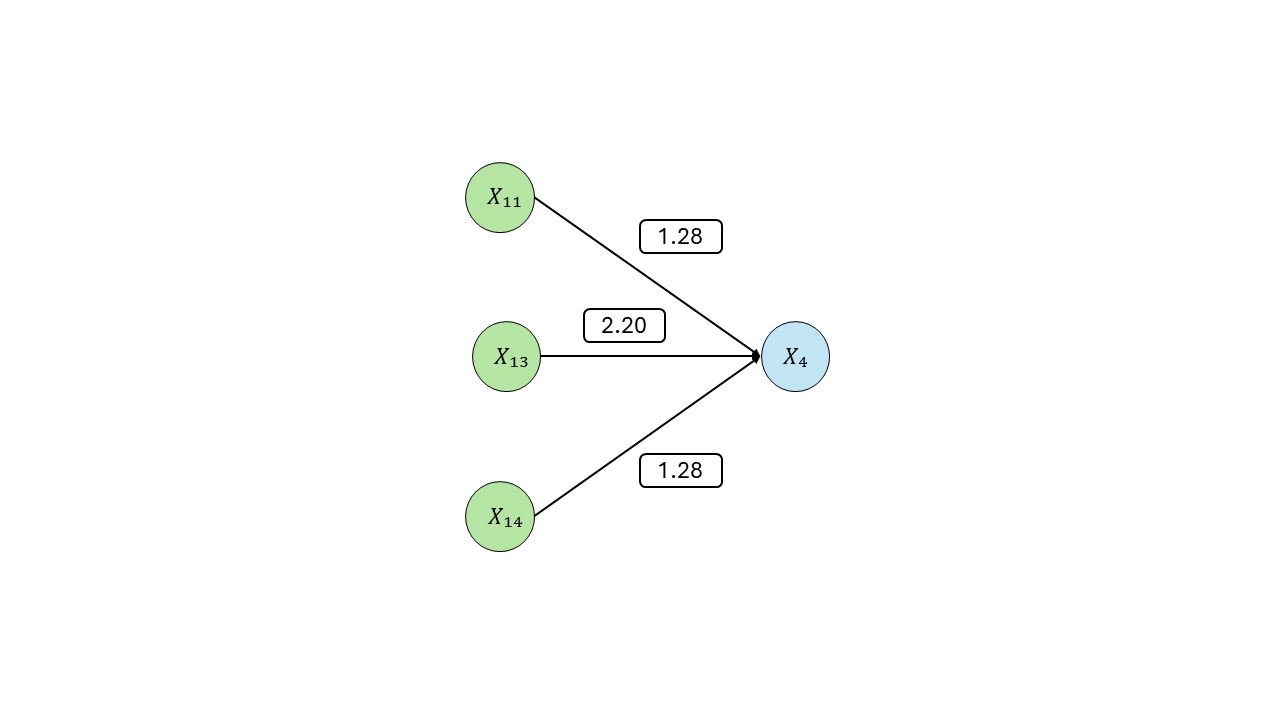}
    \caption{Causal structure for benefits per order}
    \label{fig:output2}
\end{figure}

On a more moderate level, `Sales per customer` (\(X_5\)) and `Order Item Product Price` (\(X_{12}\)) both influence `Late\_delivery\_risk` indirectly, with causal strengths of 1.10 and 1.12, respectively. These variables, while less impactful, suggest that the value of orders and customer segments might influence the priority or efficiency of deliveries, thereby affecting the overall risk of late deliveries. The combination of direct and indirect effects across these variables highlights a complex network of interactions where both logistical decisions and customer-related factors converge to influence delivery outcomes.
Fig. \eqref{fig:output2} uncovers additional patterns within the supply chain that, while not directly linked to `Late\_delivery\_risk`, offer significant business insights. The analysis reveals that both `Order Item Product Price` (\(X_{11}\)) and `Order Item Quantity` (\(X_{14}\)) moderately influence `Benefit per Order` (\(X_4\)) with a causal strength of 1.28 each, indicating that higher prices and larger quantities contribute to greater profitability per order. More notably, `Order Item Profit Ratio` (\(X_{13}\)) has a stronger influence, with a causal strength of 2.20, underscoring the critical role that profit margins play in driving overall order profitability. These findings highlight key factors in maximizing financial performance within the supply chain, suggesting that optimizing pricing strategies and profit margins could yield significant benefits. While these insights are not directly related to the root causes of late deliveries, they represent valuable areas for further research and optimization in supply chain management.

\section{Discussion and Business Impact \label{sec:4}}
We conclude that the application of causal discovery with reinforcement learning in the context of supply chain management, specifically for identifying the root causes of late delivery risks, yields several potential business insights and actionable strategies:
   \begin{itemize}
       \item \textbf{(Improving Shipping Strategies)} Our analysis has revealed that the choice of shipping mode (\(X_{16}\)) has a direct and substantial impact on the actual days for shipping (\(X_2\)), which in turn significantly influences the risk of late deliveries (\(X_7\)). The causal strength of 9.24 between shipping mode and actual shipping days underscores the critical importance of selecting the appropriate shipping method to minimize delays. By implementing our causal discovery approach, businesses could optimize their shipping strategies by prioritizing modes that reduce delivery times, thus mitigating the risk of late deliveries. This optimization can lead to improved customer satisfaction and potentially lower operational costs.
       \item \textbf{(Enhancing Delivery Status Monitoring)} The direct relationship between delivery status (\(X_6\)) and late delivery risk (\(X_7\)) with a causal strength of 10.66 highlights the need for vigilant monitoring and management of delivery processes. Our method enables businesses to identify which statuses are most predictive of delays, allowing for targeted interventions to address these bottlenecks. By improving real-time monitoring and management of delivery statuses, companies can proactively reduce the occurrence of late deliveries, ensuring more reliable service and enhancing overall supply chain efficiency.
       \item \textbf{(Strategic Pricing and Inventory Decisions)} Beyond the immediate concerns of delivery times, our analysis also identified that both `Order Item Profit Ratio` (\(X_{13}\)) and `Order Item Product Price` (\(X_{11}\)) influence the `Benefit per Order` (\(X_4\)). Although not directly related to delivery risks, these findings have implications for profitability. The causal strength of 2.20 for the profit ratio suggests that optimizing profit margins is key to maximizing order-level benefits. By using our causal discovery method, businesses can refine their pricing strategies and inventory decisions to enhance profitability. This strategic insight can lead to more informed decision-making and better financial performance across the supply chain.
   \end{itemize}

Our approach identifies the key drivers of late deliveries but also offers a broader understanding of the factors that influence overall supply chain performance. By integrating these insights into daily operations, businesses can achieve both immediate and long-term improvements in efficiency, customer satisfaction, and profitability. Future research could further explore these relationships to refine and extend these strategies, ensuring that supply chain operations remain robust and adaptive in the face of evolving challenges.

\section{Conclusions and Future work \label{sec:5}}
In conclusion, the integration of causal discovery with reinforcement learning offers a powerful tool for root cause attribution in complex supply chain environments. Our approach successfully identifies the primary factors contributing to late delivery risks, such as shipping mode and delivery status, and provides a framework for implementing targeted interventions to mitigate these risks. The ability to accurately map and understand the causal relationships within supply chains enables businesses to make informed decisions that enhance operational efficiency and customer satisfaction.

For future work, there are several avenues for future research and improvement. First, the uncertainty of calculation of causal strength can be further considered to ask questions, such as, how confident we are sure that there is a strong causal relationship between $X_i$ and $X_j$. Then, the model could be fine-tuned to handle different types of data or to adjust for varying supply chain conditions, increasing its robustness across different industries such as economics, healthcare, or manufacturing. Additionally, experimenting with different scoring functions and hyperparameters in the reinforcement learning process could further enhance the accuracy of causal discovery. The methodology could also be extended to include multi-modal data inputs or adapted to real-time data streams, providing dynamic and responsive insights that evolve with changing conditions in the supply chain. Such advancements would broaden the applicability of our approach, making it a versatile tool for optimizing performance in various domains.

\section{Acknowledgement}
ChatGPT was employed to refine the wording of some sentences in the data description subsection of Section 3.1.1 to ensure better readability and precision. The authors take full responsibility for the content of this manuscript, including sections enhanced by AI assistance. 
\bibliography{references}
\bibliographystyle{apalike}

\appendices
\section[\appendixname~\thesection]{Variable Description and Abbreviations}\label{table:des}

\begin{table}[H]
\centering
\begin{tabular}{|l|p{7.5cm}|}
\hline
\textbf{Variable} & \textbf{Description} \\ \hline
Type &   Type of transaction made \\ \hline
Days for shipping (real) &   Actual shipping days of the purchased product \\ \hline
Days for shipment (scheduled) &   Days of scheduled delivery of the purchased product \\ \hline
Benefit per order &   Earnings per order placed \\ \hline
Sales per customer &   Total sales per customer made per customer \\ \hline
Delivery Status &   Delivery status of orders: Advance shipping or not \\ \hline
Late\_delivery\_risk &   Categorical variable that indicates if sending the product would cause a late delivery\\ \hline
Customer Segment &   Business segment of the customer \\ \hline
Latitude &   Latitude coordinates of the purchase \\ \hline
Longitude &   Longitude coordinates of the purchase \\ \hline
Order Item Discount Rate &   Order item discount percentage \\ \hline
Order Item Product Price &   Price of products without discount \\ \hline
Order Item Profit Ratio &   Order Item Profit Ratio \\ \hline
Order Item Quantity &   Number of products per order \\ \hline
Order Status &   Order Status : COMPLETE , PENDING , CLOSED , etc. \\ \hline
Shipping Mode &   The following shipping modes are presented :  First Class, Second Class, Same Day, etc. \\ \hline
\end{tabular}
\caption{Variables Description}
\end{table}

\noindent 
\begin{tabular}{@{}ll}
IIE & Inverse information entropy\\
BIC& Bayesian information criterion\\
DoS& Days of shipment\\
 RL&Reinforcement Learning\\
 EDA&Exploratory data analysis\\
\end{tabular}


\end{document}